\documentclass[letterpaper]{article}
\usepackage{iccc}

\usepackage{times}
\usepackage{helvet}
\usepackage{courier}
\usepackage{graphicx}
\usepackage{enumitem}
\title{Shimon the Rapper: \\
A Real-Time System for Human-Robot Interactive Rap Battles}

\author{Richard Savery \and Lisa Zahray \and Gil Weinberg\\
Georgia Tech Center for Music Technology\\
Atlanta, USA\\
rsavery3, lzahray3, gilw @gatech.edu\\}

\begin{document}
\maketitle
\begin{abstract}
\begin{quote}

We present a system for real-time lyrical improvisation between a human and a robot in the style of hip hop. Our system takes vocal input from a human rapper, analyzes the semantic meaning, and generates a response that is rapped back by a robot over a musical groove. Previous work with real-time interactive music systems has largely focused on instrumental output, and vocal interactions with robots have been explored, but not in a musical context. Our generative system includes custom methods for censorship, voice, rhythm, rhyming and a novel deep learning pipeline based on phoneme embeddings. The rap performances are accompanied by synchronized robotic gestures and mouth movements. Key technical challenges that were overcome in the system are developing rhymes, performing with low-latency and dataset censorship.
We evaluated several aspects of the system through a survey of videos and sample text output. Analysis of comments showed that the overall perception of the system was positive. The model trained on our hip hop dataset was rated significantly higher than our metal dataset in coherence, rhyme quality, and enjoyment. Participants preferred outputs generated by a given input phrase over outputs generated from unknown keywords, indicating that the system successfully relates its output to its input.

\end{quote}
\end{abstract}

\section{Introduction}
Interactive music systems have largely focused on generating instrumental music. Lyric generation and singing synthesis have been explored, but past research did not focus on vocal musical response to human input in real-time. The field of robotic musicianship uses embodiment to improve the relationship between humans and AI, inspiring humans in new creative ways. We combine these fields to create an interactive robotic system that improvises lyrically with a human in real-time. We select hip hop as a genre well-suited toward real-time improvisation, due to art forms like freestyle rapping and battle rap.

\begin{figure} [h]
    \centering
    \includegraphics[width=7cm]{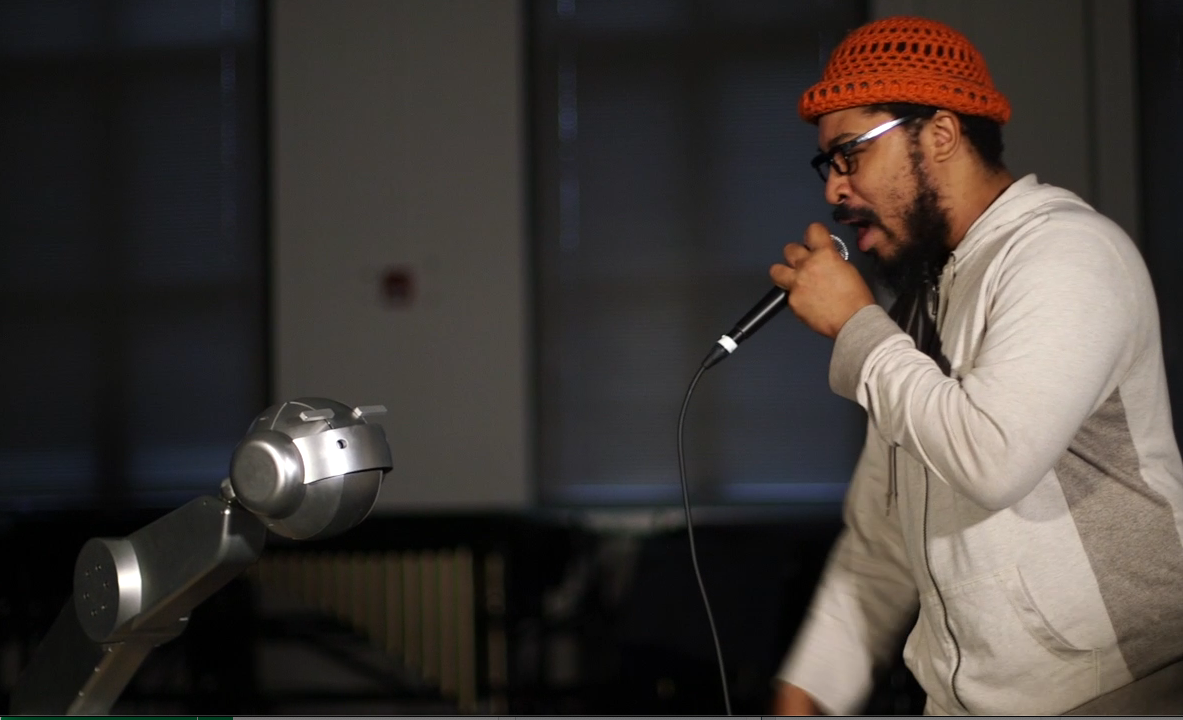}
    \caption{Shimon interacting with rapper Dashill Smith}
    \label{fig:dast}
\end{figure}
Shimon, seen in the left of Figure \ref{fig:dast}, is a marimba-playing robot who has recently been redesigned to have singing capabilities. Shimon collaborates with humans to write the lyrics to his own songs, taking keywords as input. However, in previous work, the lyric generation, voice synthesis, and gestures were not generated in real-time, and did not react to a human voice live on-stage. Our goal with this project was to allow Shimon to respond to a rapper in real-time with computer-generated rhyming lyrics, voice, rhythm, and gestures. We aim to provide the experience of a rap battle between a human and a robot, with the intention of inspiring the human rapper with machine-driven responses that are unlikely to be generated by humans.


This paper includes a technical overview of each sub-task of the system, beginning with analyzing voice input from a human rapper and ending with generating voice and synchronized gesture output by the robot. Throughout the paper, we discuss the key challenges faced during development and how they were overcome. Several examples of generated rap lyrics, rhyme analysis, and rhythm are provided. We also include a system evaluation using both quantitative and qualitative metrics. We analyze the overall perception of the system, various quality metrics of the output, and the system's success at generating output to match its input. Videos examples of the system are available here \footnote{www.richardsavery.com/shimonraps
}. To our knowledge, this is the first system with a full working pipeline of vocal input from a rapper to vocal rap output in time to a beat.


\section{Related Work}
\subsection{Generative and Interactive Music and Hip Hop}
Computerized generative music systems have been widely explored from early systems in the 1950's \cite{hiller1968music} to modern deep learning based systems \cite{briot2017deep}, tending to focus on Western Classical music and Jazz. Stylistically closer to rap generation is Algorave, where algorithms are used to create electronic dance music \cite{savery2018interactive}, although its emphasis is instrumental music. Hip hop, however, has many unique stylistic features that distinguish it from other genres and present new challenges for generative systems. Linguistically hip hop is far extended from other poetic traditions or other music forms and uses a `highly intertextual' form that `demonstrates multilayered poetic complexity'\cite{alim2003some}. Lyric delivery is commonly referred to as flow and - among many unique features - includes distinct approaches to meter, beat division, and rhyme placement \cite{condit2017mcflow,Robertthesis}.

\subsection{Lyric Generation and Voice Synthesis}
While computerized music generation has a long established history, lyric generation has only recently begun to receive attention. Past systems have focused on lyric generation based only on text without considering a musical melody, such as the Korean language lyric generator in \cite{son2019korean}. Other efforts have focused on fitting lyrics to an existing melody in Tamil \cite{ramakrishnan2010alternate}, or for Jazz \cite{watanabe2018melody}. Rap lyric generation has also been partly addressed in past work, although it has focused on hip hop as a standard natural language generation task \cite{karsdorp2019keepin}, instead of a focus on hip hop's unique aesthetic. These efforts also do not focus on real-time interaction, and often have a much more confined scope, such as generating text similar to 14 unique rappers \cite{potash2015ghostwriter}.

State of the art results in singing synthesis have recently been achieved by deep learning based systems \cite{blaauw2017neural}, building on developments made by WaveNet\cite{oord2016wavenet}. These systems, however, are computationally expensive to train and far too slow to be used for real-time generation. Comparative models that work in real-time using concatenation, such as Vocaloid \cite{kenmochi2007vocaloid}, have not matched results from offline models. Singing and voice synthesis have both been used extensively in robotic systems, such as the generation of robotic vocal prosody \cite{savery2019finding,savery2019establishing}. To our knowledge, no vocal synthesis model has focused on rap. Additionally, we believe no system has attempted to both generate lyrics and synthesize the results in real-time for a robot to interact with a human performer.

\section{System Overview}
\begin{figure} [h]
    \centering
    \includegraphics[width=8cm]{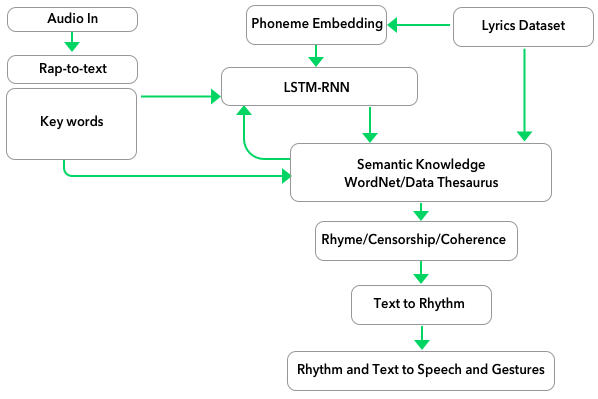}
    \caption{System Overview}
    \label{fig:systemoverview}
\end{figure}

In the following section we present an overview of the system design of Shimon the Rapper. In particular we focus on three key challenges:

\begin{enumerate}
    \item Latency and processing for real-time applications
    \item Developing internal rhymes and rhythm
    \item Approaches to censorship
\end{enumerate}

The system is written in Python, with some interfaces to MaxMSP for easy audio processing and access to external plugins for processing on the generated voice. The standard interchange for the system involves a rapper free-styling over a loop for an undetermined amount of time followed by Shimon responding. The loop can be any musical material that has a set tempo.

\subsection{Audio Analysis}
Audio from the rapper is recorded through a standard audio interface connected to an AT803 Omnidirectional Condenser Lavalier. MaxMSP is used to record the audio, breaking the rappers incoming lyrics into smaller segments. The incoming audio is chunked with a simple volume threshold, with gaps in the lyrics separated when the length of time below the threshold is over 300 milliseconds. As each clip is chunked in real-time it is written to a wav file, and a UDP message is sent from MaxMSP to Python with the file name.

Python then calls Google's Speech to Text on each segment. While using an online speech-to-text system does add some latency, chunking means there is no latency cost above the last sample sent. By keeping chunks short, the added processing time averages around 1.5 seconds of latency. We also experimented widely with offline options, testing each API linked with Speech Recognition\footnote{https://pypi.org/project/SpeechRecognition/}. However, we found Google Cloud Speech API to be the most reliable considering the background noise often present in live performances. We can opt to detect the end of the rapper's phrase by either waiting for a silence of over 700 ms or a preset number of musical bars to pass. However, to decrease latency we often end the speech to text detection at random locations, allowing Shimon to start rapping, signifying to the human that their turn is over..


\subsection{Text Analysis}
Keywords are identified from the text once it has been extracted from the audio. We implemented the TextRank algorithm \cite{mihalcea2004textrank} to categorize keywords. TextRank is a graph-based model used to rank the importance of text. In our implementation, text of up to 100 words is always processed in under 10 milliseconds. We then generate a list of synonyms and antonyms from wordnet \cite{oram2001wordnet} for each keyword.  Sentiment analysis is also implemented, as well as a system that categorizes which rapper from the dataset the incoming text is most similar to. We do not currently use these functionalities, as we have not found them helpful for the generation process.

\subsection{Dataset}
During lyric generation we primarily alternate between two custom-created datasets, switching between models in real-time. These data sets were created through a custom lyric web scraper: Verse Scraper \footnote{https://github.com/RFirstman/versescraper}. This tool was created for two main reasons. Firstly, standard datasets group all lyrics for a song together, whereas  hip hop commonly uses multiple rappers on the same track. We wanted to be able to associate lyrics with individual artists. The second benefit of our scraper is high level of customization in dataset creation, allowing us to create datasets from certain years, subsets of an artist catalogue, and other custom metrics.
For our deep learning system we use either a hip hop dataset containing 25,000 songs or a metal dataset containing 15,000 songs. We were interested in comparing these two datasets to see how well metal music lyrics transfer to the hip hop genre.

\subsection{Phoneme Embedding and LSTM-RNN}
The primary novel element in our deep learning system is the use of a phoneme embedding layer. Phonemes are the second smallest layer of word vocalization, distinguishing how words are pronounced. Groups of phonemes create each syllable, and syllables create words. Phoneme embedding has been very rarely used in generative systems with one of the only uses being in a speech recognition system\cite{yenigalla2018speech} or  occasionally in speech synthesis \cite{li2016phoneme}. In purely text-based systems, preliminary work has shown that phoneme vector spaces contain distinctive feature contrasts to word embeddings \cite{silfverberg2018sound}. We contend that due to the unique linguistic properties of hip hop, phoneme embeddings offer a promising approach for a generative system. These properties are the unique relationship between words, built on a preference for rhyme from phonemes, over common semantic meaning. These rhymes also occur at any point in the lyrics, not only at the end of lines. Additionally, hip hop flow uniquely relies on phonemes \cite{edwards2013rap} and often contains non-standard word variations and intentional variations in pronunciation to achieve flow.

The primary challenge of this approach was creating a dataset of phonemes. Mappings between the spelling of words and their phonemes are not always consistent. No extensive dataset currently exists of lyrics to phonemes, leading us to create a conversion process. While such conversion systems do exist, we found no system that can capture all the dialects used in hip hop. Our process begins by using CMU Pronouncing Dictionary\footnote{http://www.speech.cs.cmu.edu/cgi-bin/cmudict} which is based on the ARPABET phonetic transcriptions. When words are not found in the dictionary, we attempt to break the word up into the most likely phoneme subsets by searching through the dictionary for subsets of phonemes that fit the word. After phoneme subsets are found for the word, that word is then added to the dictionary so that all repeats of a word are treated the same.

With phonemes as the embedding layer, we can use a relatively standard deep learning model. While state of the art models such as GPT-2\cite{radford2019better} are based on Transformer with attention layers, we used an encoder/decoder RNN-LSTM, as we aimed for real-time generation. Many of the advantages of larger models are for improved long term structure, which is not required for short phrases such as the ones we are creating.

We first generate many lines of text with the model. We then automatically choose phrases from all the generations that utilize either a keyword, or a synonym or antonym from the keywords. This process allows us to combine multiple generations and meanings, while still placing an emphasis on internal rhymes created through deep learning, and line by line rhymes through rhyme detection.

\subsection{Censorship}
Censorship of the output was a significant consideration and design challenge for the system. We first created a list of 28 words that would not be appropriate for the system to output. For some words this list included multiple spelling variations. After creation, the list was encoded with ROT13, to allow us to more comfortably share the code.

We considered multiple approaches to censorship, aiming to balance maintaining authenticity of the dataset, while meeting language requirements. In original tests we considered excluding from the hip hop dataset any song that contained one of the words in our list of censored words. This reduced the data size from 25,000 songs, to 7,000. To counter this we considered removing lines or whole verses containing certain words. Given hip hop's reliance on flow, this approach proved ineffective as it seemed to significantly alter the data set. Likewise we considered replacing offending words with a substitute, but again, due to subtle elements impacting rhythm and flow this was deemed as an inappropriate method. To maintain authenticity we decided to keep the original dataset and instead censor phrases by post processing created material. After creation we discard any generation that includes a filtered word and create a new generation as a replacement. While this does add extra processing time into the system, we found it a worthwhile trade off.

\begin{figure} [h]
    \centering
    \includegraphics[width=6.2cm]{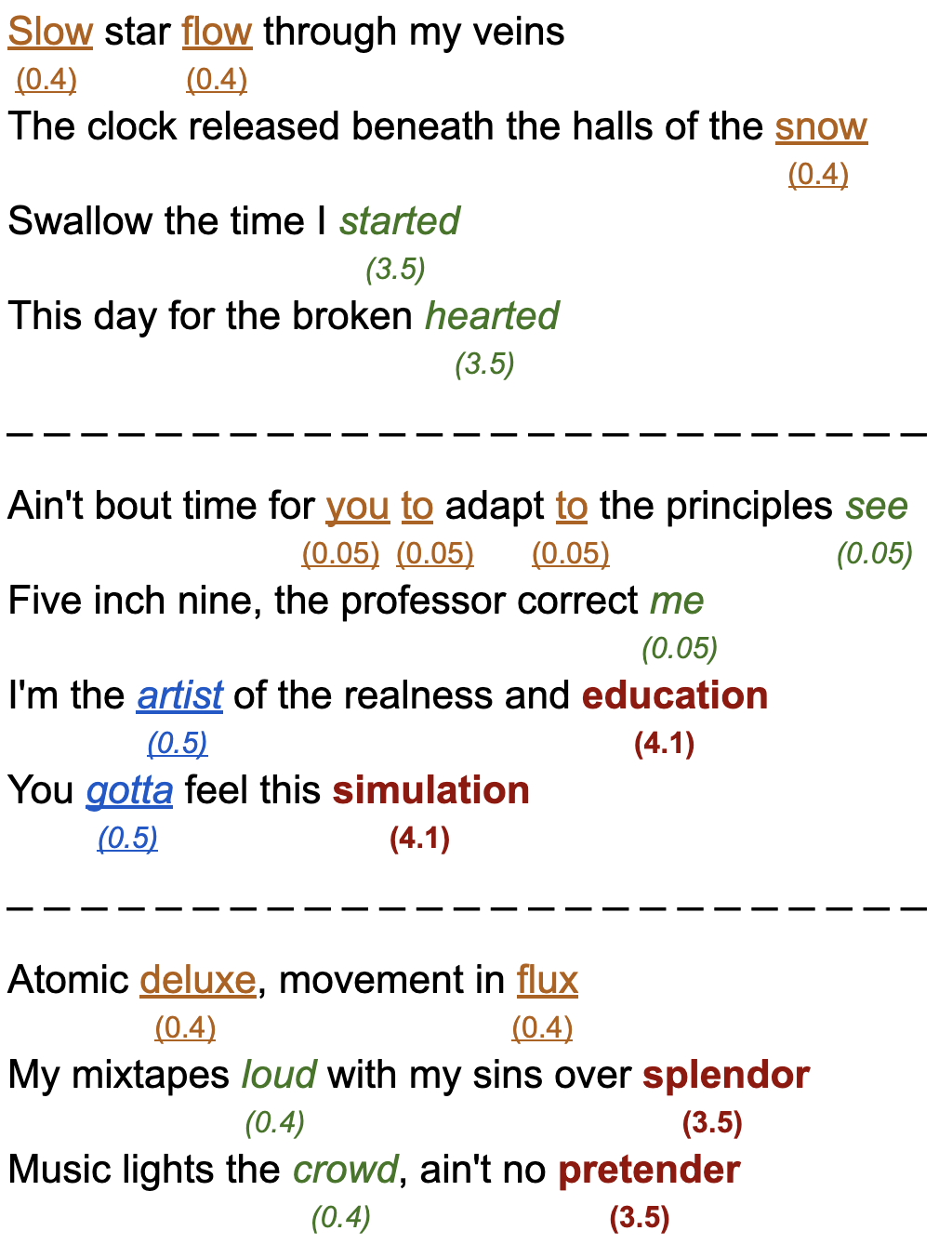}
    \caption{Generated lines with rhyme detection and scoring}
    \label{fig:rhyme}
\end{figure}

\subsection{Rhyme Detection and Choices}
The phoneme embedding naturally generates lines of text containing internal rhymes. We automatically select which generated lines to use based on the quantity and quality of internal rhymes of each line, as well as which lines rhyme best with each other. We originally tried using existing rhyme libraries, but found that they were too slow when iterating through a large number of words. We created our own implementation for scoring rhymes that runs in a few hundredths of a second on large numbers of phrases.

We are interested in detecting and scoring two types of rhyme: perfect rhymes, where vowels and consonants match, and slant rhymes, where words have similar but not identical sounds. For our system, we specify slant rhymes as vowels that match, but consonants that may not match. We create dictionaries for each line, recording phoneme patterns and their frequency of occurrence within the line. Our first dictionary type is for perfect rhymes, which records the last one, two, and three-syllable sequences (vowels and subsequent consonants) of each word. The second dictionary is for slant rhymes, which records the last two and three-vowel sequences of each word. Finally, we create a dictionary excluding words that are only 3 or fewer phonemes long. This allows us to give a lower score to words that rhyme perfectly but are very short (such as "to" and "do"). For all dictionaries, a sequence of vowels or syllables is only added if it contains at least one stressed vowel.


We first use these dictionaries to select the line with the highest internal rhyme score. Each type of phoneme pattern is assigned a score, where perfect rhymes and a higher number of  matching syllables are scored higher than slant rhymes and fewer matching syllables. These scores are summed according to the number of instances of each detected rhyme. We do not count multiple occurrences of the same word as any type of rhyme.

To select each subsequent line, we find the line that rhymes best with the previously selected line. We do this by calculating the rhyme score for phoneme patterns that are present in both lines. After all lines are selected, we assign each rhyme group a unique number to identify which words rhyme with which other words. This information is used in the next steps of the pipeline. Figure \ref{fig:rhyme} shows examples of generated lines that were selected using the rhyme scoring, along with the calculated rhyme scores.

\begin{figure} [t]
    \centering
    \includegraphics[width=8cm]{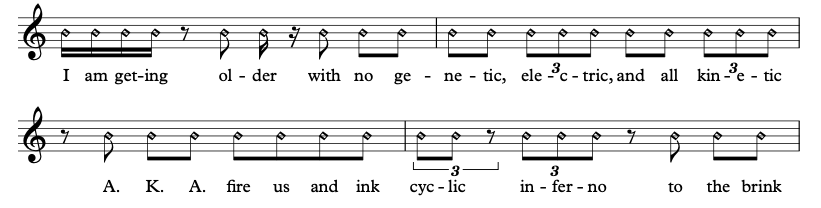}
    \caption{Generated response to keywords electric and genetic}
    \label{fig:generated response}
\end{figure}
\subsection{Text to Rhythm}
We next generate rhythms for the generated text, with each syllable assigned a time. We designed a rule based system for rhythmic generation based on concepts presented by Edwards \cite{edwards2009rap,edwards2013rap}.  Edwards compiled a collection of interviews with over one-hundred leading hip hop artists discussing their approach for  flow and rhythm. Based on these interviews, we designed a rhythm generation that is able to map to any tempo, although has been primarily used for tempos ranging from 80 to 160 beats per minute.

Rhythmic generation focuses on emphasizing rhyming words. Emphasis is added to words by expanding the length of the word beyond non-rhymes and by placing different lengths of silence after each rhyming word. Keywords with more than one syllable are set as quarter-note triplets, allowing them to stand out from non-keywords without interrupting the flow. All non-rhyming words are set as eighth notes. In experiments we also applied similar rules to nouns or other word types, but found this was not represented in texts and was not well received internally. Figure \ref{fig:generated response} shows an example of a generated phrase with its corresponding rhythm.

\subsection{Rhythm to Voice} Shimon's voice is generated by modifying the output of Google's text-to-speech system. Speech Synthesis Markup Language (SSML) provides options for changing vocal prosody in text-to-speech systems. We use SSML to emphasize any word that rhymes with at least one other word. We additionally pitch-shift all matching rhymes by the same amount, allowing for either upwards or downwards pitch shifts. This method, which is common in hip hop \cite{Robertthesis} can help the listener to notice which words rhyme with each other .

In order to place the words at the correct time according to the generated rhythm, we originally tried running text-to-speech separately on each word. However, we found that this took too much time and could result in an unnatural cadence when the words were strung together. Therefore, in order to quickly generate the audio, we run text-to-speech once on the entire sentence with added breaks between each word using SSML. We then split the resulting audio file into the non-silent segments to separate the words.

We found that the endings of individual words are frequently cut off upon generation, due to the way words flow together in natural conversation. This made it more difficult to generate arbitrary rhythms from the words, as a long break after a cut-off word could sound odd and difficult to understand. To address this issue, as well as to better match the generated rhythm for multi-syllable words, we time-stretch the words to flow more naturally into each other.

We align each synthesized word to start at the time given by the rhythm generation. We end each word's audio on whichever occurs first: the start time of the following word, or a tempo-dependent offset after the start time of the word's last syllable. This helps words that are close together flow into each other more naturally, while also stretching words that precede longer gaps to mitigate any cut-off endings.

While our generated rhythm provides start times for each syllable in a word, we only modify word start and end times when generating the audio. A more complex audio analysis could have allowed for alignment of each syllable. However, we chose not to do this to increase system robustness, and to maintain the original timings produced by the text-to-speech system to preserve naturalness. Finally, we compress and filter the output to improve the audio quality, using commercial audio plugins\footnote{https://polyversemusic.com/products/manipulator/}\footnote{https://slatedigital.com/}. This also raises the overall pitch of Shimon's voice, producing a unique and cute voice timbre befitting of Shimon's persona as a robotic rapper.

\subsection{Gesture Synchronization}
\begin{figure} [b]
    \centering
    \includegraphics[width=8.3cm]{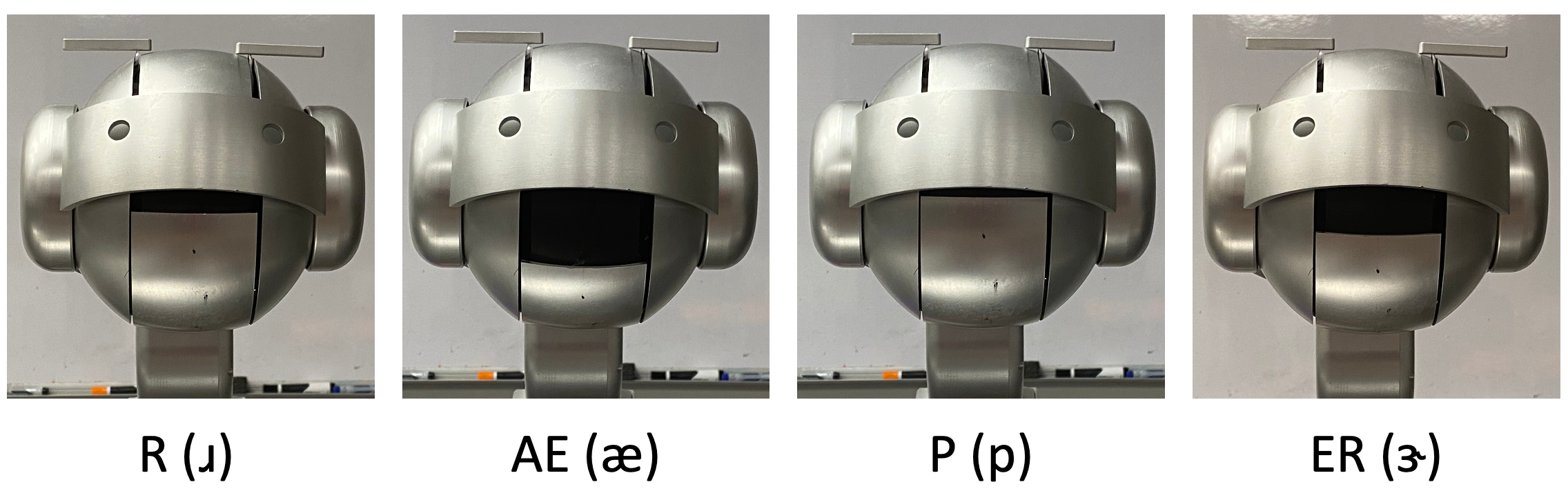}
    \caption{Example poses of Shimon's mouth for each phoneme of the word `rapper'}
    \label{fig:mouthPoses}
\end{figure}

Shimon's gesture design while rapping consist of both synchronizing his mouth movements to the audio, as well as generating head and neck movements throughout the rap battle. We create the mouth movements using each syllable's times from our rhythm generation. Similarly to the key pose approach used in \cite{tachibana2010singing}, Shimon's lip syncing linearly interpolates between phoneme-dependent positions. Some examples of these positions can be seen in Figure \ref{fig:mouthPoses}. The linear interpolation allows for smooth easing into and out of mouth poses. Consonants are given a default maximum duration. However, if a syllable's vowel duration is shorter than the default consonant duration, half of the syllable time is given to the vowel and the remaining time is evenly divided among consonants. If a word is not found in the CMU Pronouncing dictionary, it is assigned the phonemes [P, AH, P] by default.

We raise Shimon's eyebrows on rhyming words to increase emphasis. We also do this with the intention of conveying that Shimon is pleased with his generated rhymes and is challenging the human rapper interacting with him. When Shimon is listening to the rapper, we have him nod to the beat, move his head side to side on every downbeat, and move his body up and down on every other downbeat. We position his head and body so he is looking at the rapper. While performing his own rap, Shimon slowly moves side to side and up and down with the beat, keeping his head positioned so that his mouth is visible.

\subsection{Latency}
\begin{figure} [t]
    \centering
    \includegraphics[width=8.4cm]{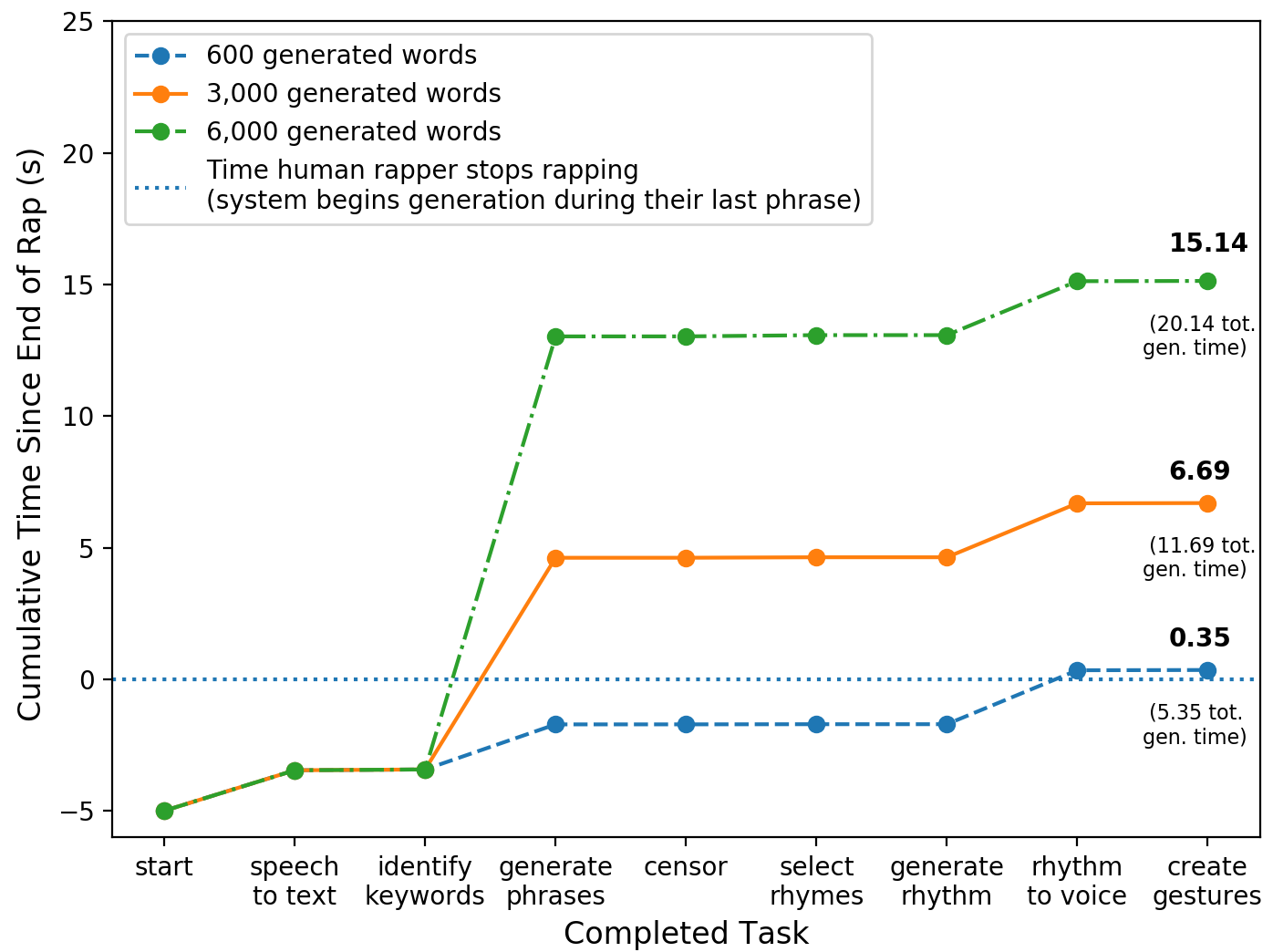}
    \caption{Average latency for each subtask}
    \label{fig:latency}
\end{figure}

Latency is a constant trade-off between time and quality. The most time-intensive tasks are rap-to-text, generating phrases to select from, and rhythm-to-voice. All other tasks are on the order of hundredths of a second or lower. Figure \ref{fig:latency} shows the time required for each sub-process, highlighting the time difference between different numbers of generated words.
We find that generating around 3,000 words is a good compromise for maintaining high quality with low latency. This number can be higher for a more powerful GPU. With these settings, the time between starting the generation and when Shimon begins rapping is approximately 11.69 seconds. However, because we can start the generation while the human rapper is still finishing, the perceived latency can be made to be lower. By ignoring the rapper's last sentence, that time can be reduced to 6.69 seconds, during which an instrumentalist can play a quick solo or Shimon can use gestures to stall while generation completes.

The pipeline makes use of two computers, one that uses a 1080 GPU to generate choices for output phrases, and another that performs all other audio and computational tasks. Two computers are used due to compatibility with MaxMSP.

\section{System Evaluation}
A broad Turing style test, as is often used \cite{agres2016evaluation}, does not make sense for this system since by definition, Shimon lyrical output does not aim to sound like a human. Likewise, computer based NLG or chatbot metrics tend to focus on features that are not easily applied to our design - such as readability and grammatical correctness - and have been shown to give significantly different results than human ratings for creative tasks\cite{novikova2017we}. There are multiple non-academic frameworks that exist to evaluate human-created hip hop and rap, however many of these tools were referenced in the creation process and would be unfairly biased towards our system \footnote{https://www.rappad.co/blueprints/faq}.

Additionally, throughout the design and development stage we regularly engaged with Atlanta rapper Dashill Smith. This involved five extended sessions where he informally analyzed and reviewed the system output. These sessions led to an iterative design process, where we would build on and alter the system based on his reactions and insights. During this stage of the project, we chose not to collect formal data from Smith, instead allowing for natural discussion and broad ideas for future directions and improvements. While evaluation by experts through interaction has been shown as an effective means to analyze interactive systems \cite{bown2015player}, we chose to frame our final evaluation around audiences' perception, enjoyment, and rating of the system, since while the system is interaction-based its ultimate use case is in musical performance to listeners. In future research we aim to engage multiple rappers with the system for evaluation.

With these challenges in mind we designed our evaluation to answer the following research questions:
\begin{enumerate}
    \item What is the perception of the system by listeners and what do subjects think about idea of a robot-human having rap battles in general?
    \item Can we create high quality stand alone hip hop?
    \begin{enumerate}[leftmargin=.75cm, label*=\arabic*.]
    \item Do our stand alone rap outputs lead to good coherence, rhythm, rhyme, quality, and enjoyment?
    \item Are there differences in these metrics when using the hip hop dataset versus the metal dataset?
    \end{enumerate}
    \item Is there a clear relationship between the system's output and its input?
\end{enumerate}

\subsection{Method}
33 participants answered survey questions about videos and text samples generated by the system. The participants were undergraduate students recruited from the Georgia Tech School of Psychology participant pool. Participants were not required to have any musical experience, however chose to participate in the experiment based on their interest in the topic. We calculated the minimum amount of time it should take participants to complete the survey, watching all videos and reading all text samples, and eliminated 6 participants who completed it in less than that amount of time. This left us with 27 remaining participants.

First, participants were introduced to the project's concept by watching a 45-second video clip of a rapper freestyle rapping back and forth with Shimon. They were asked to describe their thoughts on the footage. This data was used to answer Research Question 1. In past studies analyzing text has been shown to provide insightful information on robot perception\cite{vlachos2018public}. In generating each sample shown to participants for the remainder of the survey, we ran the model three times on its keyword input and hand-selected one response we believed to be the highest quality.

To address Research Question 2, we presented subjects with 10 randomly-ordered videos of Shimon  performing a rap with subtitles, without being shown the input to the system. To generate the raps for these videos, 5 distinct sets of keywords were used. Each keyword set generated two of the samples, one using the hip-hop dataset and the other using the metal dataset. The participants were asked to rate the coherence, rhythm, rhyme, overall quality, and overall enjoyment of each sample on a scale from 1 to 7.

To address Research Question 3, participants were given 10 randomly-ordered tasks selecting which of two text samples they preferred as a response to given input text. One of these responses was generated by the model in response to the given input, and the other response was generated by a keyword unrelated to the input text. All keywords were randomly sampled from words that occurred over 10 times in the dataset. The order in which the responses were presented was randomized as well. Within each question, the two responses were generated using the same dataset, where 5 questions used the hip-hop dataset and 5 used the metal dataset.

\subsection{Results}

\subsubsection{Perception}
\begin{figure} [t]
    \centering
    \includegraphics[width=8cm]{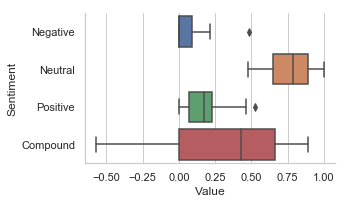}
    \caption{Sentiment of Comments}
    \label{fig:Sentiment}
\end{figure}

\begin{figure} [b]
    \centering
    \includegraphics[width=8.5cm]{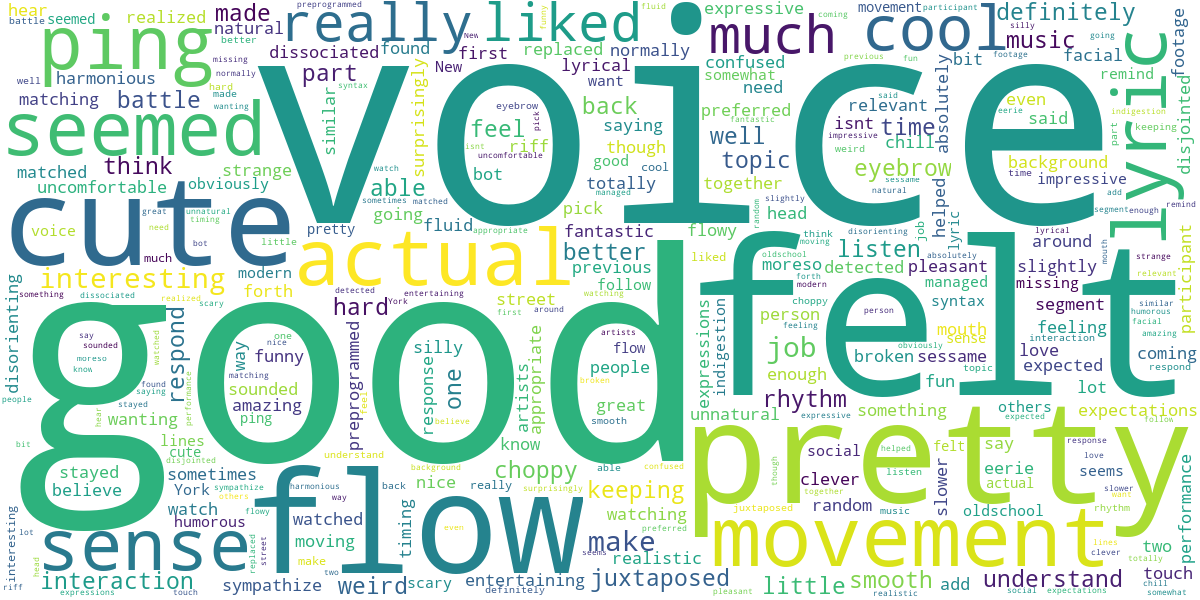}
    \caption{Word Cloud for comments}
    \label{fig:wordcloud}
\end{figure}

From the collected text responses we firstly analyzed the sentiment of each response, using the Valence Aware Dictionary and sEntiment Reasoner (VADER) \cite{hutto2014vader}. This provided us with a value for negative, positive, neutral and compound sentiment (see Fig.\ref{fig:Sentiment}. The compound sentiment is a normalized, weighted composite score between -1.0 (negative) and 1.0 (positive). The mean of the compound sentiment was 0.33, indicating an overall positive perception of the system.

Subjects' comments covered a wide range from `very expressive' and `amazing' to `the robot's voice sounded very strange when juxtaposed with the human'. A common thread was participants describing the generation as `better than expected'. We also found most comments focused on the voice with less emphasis on the lyrics. See Figure \ref{fig:wordcloud} for a word map with the most common words in subjects responses (excluding standard stopwords and the word robot).

\subsubsection{Rap Quality and Data Set Comparison}

\begin{figure} [t]
    \centering
    \includegraphics[width=8cm]{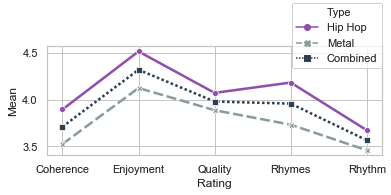}
    \caption{Means of Hip Hop and Metal Datasets}
    \label{fig:means}
\end{figure}

In each of the categories, the hip hop data set achieved a slightly higher mean (see Fig \ref{fig:means}). Comparing the hip hop and metal dataset using an independent samples t-test we found two insignificant results for rhythm (p = 0.226) and quality (p=0.225). This makes sense as rhythm is generated independently of the data set and quality should consider the system as a whole. The enjoyment was significant with hip hop being slightly favored (p = 0.046). We also found significant results in the coherence(p=0.027) and rhymes (p=0.017).

\begin{figure} [b]
    \centering
    \includegraphics[width=8cm]{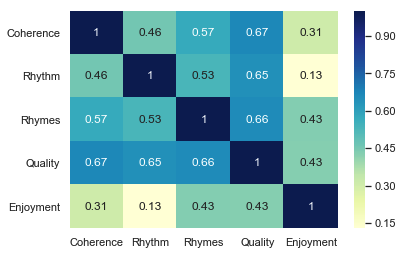}
    \caption{Correlation Matrix}
    \label{fig:corr}
\end{figure}
We found the lowest correlation between rhythm and enjoyment, while there was a strong correlation between the perceived quality and coherence, rhythm, and rhymes (see Fig \ref{fig:corr}). Importantly, we found no clear correlation between any category and the participants actual rated enjoyment of the rap, perhaps implying we need to consider other metrics for our generation system.

\subsubsection{Input and Output}
\begin{figure} [t]
    \centering
    \includegraphics[width=7.5cm]{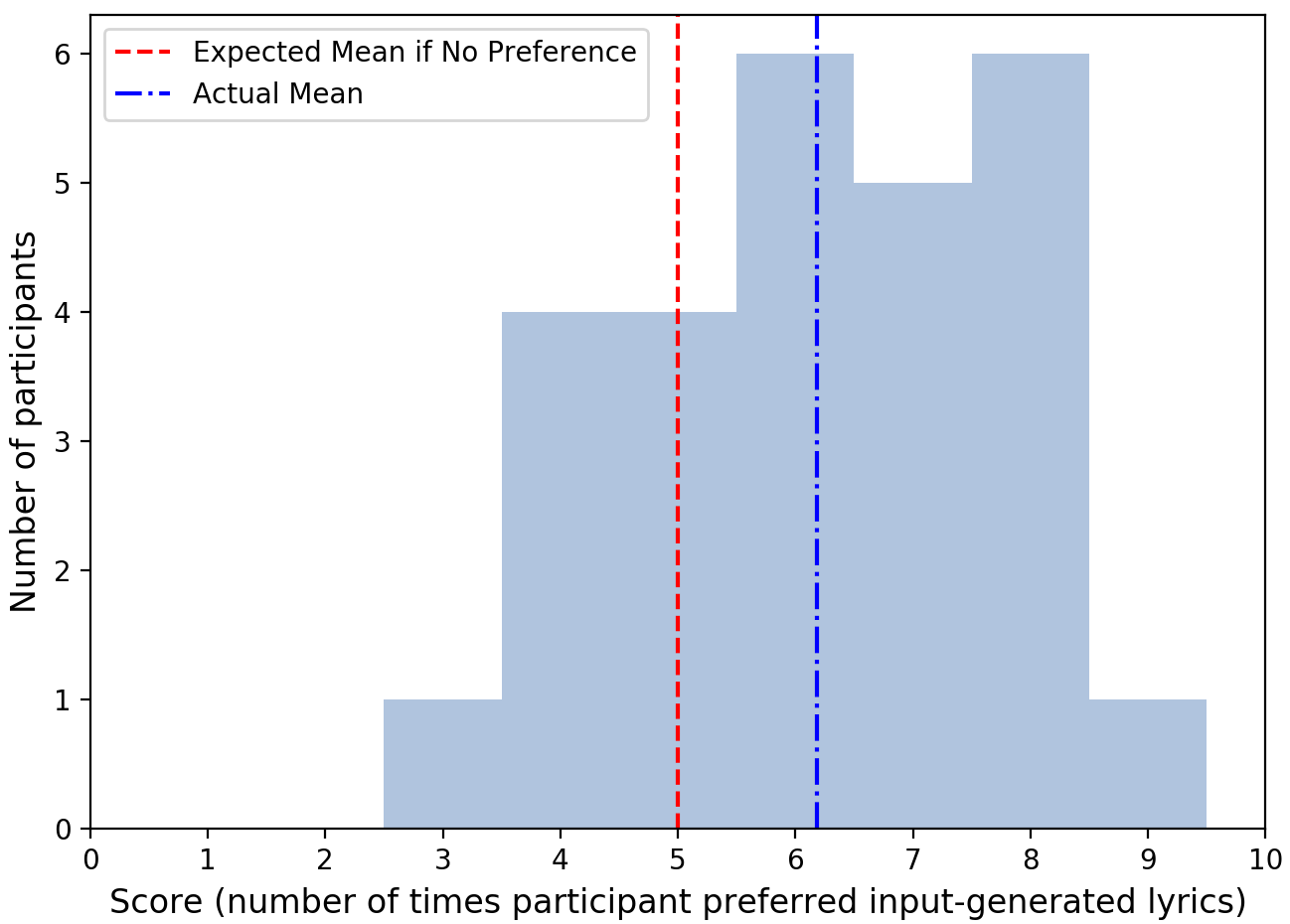}
    \caption{Histogram of participant scores when selecting preferred lyrics in response to given input}
    \label{fig:textPreference}
\end{figure}

To address Research Question 3, we evaluated whether participants preferred lyrics generated from the given input over lyrics generated from unknown, random keywords. We assigned each participant a score, defined as the number of times (out of the 10 questions) they preferred the lyrics generated by the given input over an unknown random input. We then performed a 1-tailed, 1-sample t-test on these scores, comparing against an expected mean of 5 out of 10. The p-value is 0.00033, which is less than the alpha of 0.05. The average score across all participants was 6.2 out of 10. The data support that the participants preferred lyrics generated from the given input over a random input. Figure \ref{fig:textPreference} shows the distribution of participants' scores.

\section{Discussion and Conclusion}

 In our evaluation, participants expressed overall positive sentiment regarding the perception of the system. The hip hop dataset was rated significantly higher than the metal dataset in several categories. This could be due to metal music relying more heavily on melody than hip-hop, thus being less appropriate for rap-style delivery. Interestingly, the enjoyment rating did not have high correlation with any other category, which could indicate that other evaluation metrics may be more relevant to evaluating this type of system. Our evaluation supported that participants preferred responses generated from the given input over samples generated from randomized keywords. This supports that participants recognized that the system's output related to its input. However, the number of compared samples was small, so it is possible there were other reasons for this preference.

Each task within the system has room for improvement in quality and latency. Occasionally, our rhyme detection system may miss or incorrectly identify rhymes if the word is not found in the CMU Pronouncing dictionary, or if it has multiple possible pronunciations.
More work into word pronunciation given context in a sentence would help improve both the phoneme embedding and rhyme scoring tasks.

Future work in rhythm generation could use a data-based approach, as opposed to our strictly rule-based system. MCFlow \cite{condit2017mcflow} is one example of a dataset that could be useful for this. It would also be interesting to approach rhythm generation for different styles of rap.

Currently, the head and body gestures are predetermined, with only eyebrow and mouth gestures dependent on the rap's content. Incorporating computer vision could allow for more personalized interactions, such as following the rapper as they move across the room, and potentially matching the way they move to the beat.



 The pipeline we have established allows for modifying the settings and even the overall approach to each subtask.  As we gain more feedback from rappers interacting with the system, we will continue making improvements. We hope to use this system to inspire rappers through the novel experience of an interactive rap dialogue with a robot.

%


\section{Acknowledgements}
We would like to thank Eddy Chiao, Brian Model, Jacob Meyers, Naveen Ram, Rob Firstman, and Spencer Gold for their contributions to prototyping the system.

\bibliographystyle{iccc}
\bibliography{iccc}

\end{document}